\documentclass{article}

\usepackage{arxiv}
\usepackage[utf8]{inputenc}
\usepackage[T1]{fontenc}
\usepackage{hyperref}
\usepackage{url}
\usepackage{booktabs}
\usepackage{amsfonts}
\usepackage{amsmath}
\usepackage{amssymb}
\usepackage{nicefrac}
\usepackage{microtype}
\usepackage{graphicx}
\usepackage[numbers]{natbib}
\usepackage{doi}
\usepackage{pgfplots}
\usepackage{tikz}
\usepackage{siunitx}
\usepackage{subcaption}
\usepackage{algorithm}
\usepackage{algpseudocode}
\usepackage{todonotes}
\usepackage{fontawesome5}
\usepackage{placeins}

\usetikzlibrary{arrows.meta, positioning, shapes.geometric, calc, decorations.markings, patterns, fit, backgrounds}
\usepgfplotslibrary{groupplots}
\pgfplotsset{compat=1.18}

\newcommand{\vect}[1]{\mathbf{#1}}
\newcommand{\mat}[1]{\mathbf{#1}}

\DeclareMathOperator{\logsumexp}{LogSumExp}

\newcommand\figref{Figure~\ref}
\renewcommand\eqref{Equation~\ref}
\newcommand\tabref{Table~\ref}
\newcommand*{\secref}[1]{\hyperref[{#1}]{\S\ref*{#1}}}

\title{Differentiable Satellite Constellation Configuration via Relaxed Coverage and Revisit Objectives}

\author{
    \href{https://orcid.org/0000-0002-7227-4946}{\includegraphics[scale=0.06]{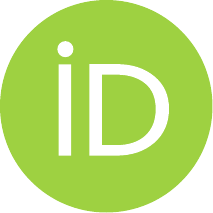}\hspace{1mm}Shreeyam Kacker} \\
	Department of Aeronautics and Astronautics\\
	MIT\\
	\texttt{shreeyam@mit.edu}
    \And
    \href{https://orcid.org/0000-0002-7791-5124}{\includegraphics[scale=0.06]{orcid.pdf}\hspace{1mm}Kerri Cahoy} \\
    Department of Aeronautics and Astronautics\\
    MIT\\
    \texttt{kcahoy@mit.edu}
}

\renewcommand{\headeright}{}
\renewcommand{\undertitle}{}
\renewcommand{\shorttitle}{Differentiable Constellation Design}

\hypersetup{
pdftitle={Differentiable Constellation Design via Relaxed Coverage and Revisit Objectives},
pdfsubject={astro-ph, cs.LG},
pdfauthor={Shreeyam Kacker, Kerri Cahoy},
pdfkeywords={satellite constellation, differentiable programming, orbit propagation, coverage optimization},
}

\begin{document}
\maketitle

\begin{abstract}
Satellite constellation design requires optimizing orbital parameters across multiple satellites to maximize mission specific metrics. For many types of mission, it is desirable to maximize coverage and minimize revisit gaps over ground targets. Existing approaches to constellation design either restrict the design space to symmetric parametric families such as Walker constellations, or rely on metaheuristic methods that require significant compute and many iterations. Gradient-based optimization has been considered intractable due to the non-differentiability of coverage and revisit metrics, which involve binary visibility indicators and discrete max operations. We introduce four continuous relaxations: soft sigmoid visibility, noisy-OR multi-satellite aggregation, leaky integrator revisit gap tracking, and LogSumExp soft-maximum, which when composed with the $\partial$SGP4 differentiable orbit propagator, yield a fully differentiable pipeline from orbital elements to mission-level objectives. We show that this scheme can recover Walker-Delta geometry from irregular initializations, and discovers elliptical Molniya-like orbits with apogee dwell over extreme latitudes from only gradients. Compared to simulated annealing (SA), genetic algorithm (GA), and differential evolution (DE) baselines, our gradient-based method recovers Walker-equivalent geometry within ${\sim}750$ evaluations, whereas the three black-box baselines plateau at with significantly worse revisit even with roughly four times the evaluation budget.
\end{abstract}

\noindent\textbf{Keywords:} satellite constellation design, differentiable programming, orbit propagation

\noindent\faGithub\hspace{0.5em}\url{https://github.com/shreeyam/differentiable_eo}

\section{Introduction}
\label{sec:introduction}

% Satellite constellation design---selecting orbital parameters for a fleet of satellites to optimize Earth observation performance---is a central problem in mission planning~\citep{wertz2011space}.
% For missions requiring responsive coverage of specific regions or targets, the design challenge is acute: the constellation must provide frequent revisit over non-uniform, mission-specific areas of interest while respecting orbital mechanics constraints.
% Applications range from disaster monitoring~\citep{paek_optimization_2019} and maritime domain awareness to agricultural surveillance and defense reconnaissance.
Designing satellite constellations is a fundamental problem in space mission design, across applications such as Earth Observation (EO), telecommunications, and navigation. The design space is high dimensional, with each spacecraft bringing up to six degrees of freedom corresponding to their orbital elements, and a wide range of often competing objectives to optimize over---for example, coverage and revisit over targets of interest, daytime illumination for optical payloads, and downlink contact time with ground stations.

Traditional approaches address this complexity in one of two ways.
Parametric families such as Walker Delta, Walker Star~\citep{walker_satellite_1984}, and Flower constellations~\citep{mortari_flower_2004} reduce the design space to a fixed set of parameters but are inherently symmetric, making them unable to capture non-uniform objectives.
Metaheuristic methods such as genetic algorithms can handle arbitrary objectives but require thousands to millions of simulator rollouts~\citep{paek_optimization_2019}.

For EO missions, the design problem has additional complexity---coverage and daytime revisit requirements may be non-uniform across ground targets, with additional constraints on data downlink. The combinatorial nature of the problem makes metaheuristic methods computationally expensive~\citep{paek_optimization_2019}, with computational cost growing rapidly with the number of satellites and the number of free parameters.

Gradient-based optimization has been previously attempted for constellation design, but with limited success. \citet{paek_optimization_2019} found the problem poorly conditioned for gradient-based methods, with difficulty even achieving convergence. Previous work has been limited by the non-differentiability of the computational graph, requiring finite difference approximations of gradients~\citep{paek_optimization_2019}. Recent work by \citet{acciarini_closing_2025} addresses this gap by developing $\partial$SGP4, an end-to-end differentiable SGP4 orbit propagator. However, the coverage and revisit metrics themselves remain non-differentiable, due to requiring binary visibility indicators, Boolean OR operations, and max functions.

\begin{figure*}[h]
    \centering
    \includegraphics[width=.85\linewidth]{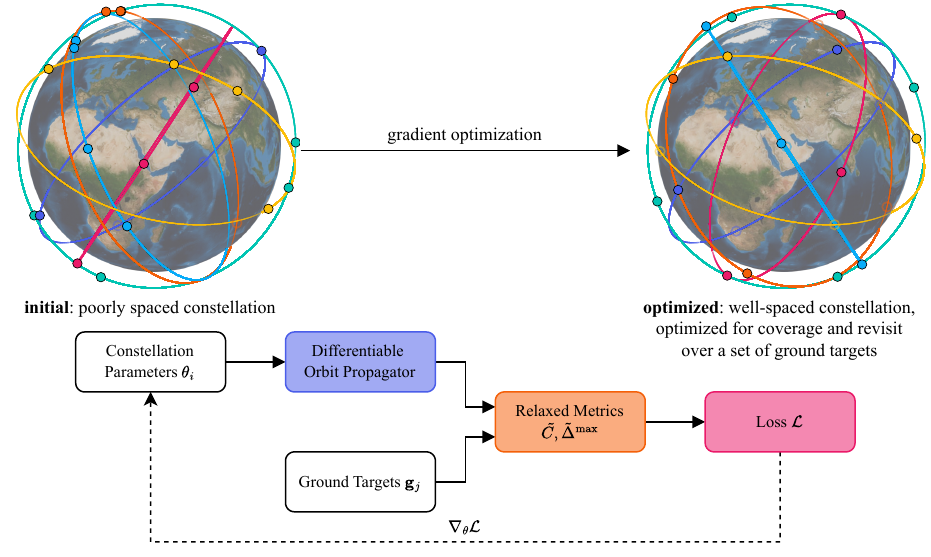}
    \caption{%
        Overview of end-to-end differentiable constellation optimization.
        \textbf{Top:} Starting from a constellation with irregularly spaced orbital planes and satellites, gradient optimization through the differentiable pipeline recovers a well-spaced constellation optimized for both coverage and revisit for a set of ground points.
        \textbf{Bottom:} Orbital parameters $\boldsymbol{\theta}_i$ are propagated through a differentiable orbit propagator~\citep{acciarini_closing_2025} and
        evaluated against a set of ground targets $\vect{g}_j$ to produce relaxed coverage and revisit metrics ($\tilde{C}$, $\tilde{\Delta}^{\max}$), which form a
        differentiable loss $\mathcal{L}$. Gradients
        $\nabla_{\boldsymbol{\theta}}\mathcal{L}$ flow back
        to the orbital parameters via reverse-mode automatic differentiation and are consumed by the optimizer to produce the next iterate.%
    }
    \label{fig:pipeline}
\end{figure*}

In this work we focus on constellation configuration: given a fixed fleet of $N$ satellites, we assign orbital elements to best meet mission-level coverage and revisit objectives. We make the following contributions:

\begin{enumerate}
    \item \textbf{Four continuous relaxations} that make coverage and revisit objectives differentiable: a soft sigmoid for visibility, a noisy-OR for multi-satellite aggregation, a leaky integrator for revisit gap tracking, and a LogSumExp soft-maximum for worst-case revisit (\figref{fig:relaxations}).
    \item \textbf{An end-to-end differentiable pipeline} from orbital elements to mission-level coverage metrics, composed with $\partial$SGP4~\citep{acciarini_closing_2025} and optimized with the AdamW optimizer (\figref{fig:pipeline}).
    \item \textbf{Demonstration on weighted target optimization}, recovering Molniya-like orbits for coverage over extreme latitudes (\secref{sec:experiments}).
\end{enumerate}

\paragraph{Paper organization.}
We review prior work on constellation design and differentiable physics simulation (\secref{sec:related_work}) before formulating the exact problem and introducing the relaxations required to make constellation configuration differentiable (\secref{sec:method}). We then conduct experiments on a toy problem, Walker-delta recovery from an irregular initialization, and weighted regional target optimization (\secref{sec:experiments}), benchmarking against tuned SA/GA/DE baselines with ablations isolating each design choice. We close (\secref{sec:conclusion}) with the scope of the configuration problem and the mission-driven constraints that fit naturally into the same method.

\section{Related Work}
\label{sec:related_work}

\subsection{Constellation Design and Optimization}

Parametric constellations significantly reduce the number of design parameters, with the Walker Delta pattern \citep{walker_satellite_1984} requiring only the number of planes, satellites per plane, and inclination. Flower constellations extend these ideas by using elliptical orbits to form specific ``harmonics'' in revisit rates to produce repeating ground tracks \citep{mortari_flower_2004}, allowing for directed coverage over a specific area. These parametric families are efficient but inherently symmetric, and cannot be tailored to non-uniform or mission-specific coverage requirements without breaking the symmetry assumptions that make them tractable.

For non-parametric design, previous work has relied primarily on metaheuristic methods.
\citet{paek_optimization_2019} uses genetic algorithms and simulated annealing to reconfigure constellations, optimizing for global coverage and regional revisit over five design variables, requiring approximately 2500 total evaluations through computationally expensive STK orbit simulations.
They also attempted gradient-based optimization via finite differences and found it ``extremely poorly conditioned,'' with convergence rarely achieved.
In a survey of 144 papers on constellation design, \citet{choo_survey_2024} finds that gradient methods appear as a design approach in only two of the surveyed articles, making it among the least common methods---ranking far behind genetic algorithms, simulated annealing, and analytical approaches.
Neither of the two entries actually uses analytic gradients.
The first, \citet{fraire_sparse_2020}, is labeled a gradient method but in fact uses a greedy neighbor search over discrete parameters (number of satellites, planes, inclination) with no gradients computed.
The second, \citet{abdelkhalik_optimization_2011}, pairs a genetic algorithm with a ``second-order gradient'' refinement step to cover ground sites under $J_2$ perturbations and explicitly notes that the fitness landscape contains numerous local minima that cause classical optimization methods to fail---but the gradient refinement is applied to a single-orbit subproblem rather than the multi-satellite design variables, so the overall design loop remains heuristic.
Separately, \citet{williams_rogers_optimal_2026} formulate constellation configuration as a collection of mixed-integer linear programs that select which satellites to activate from a pre-enumerated set of orbital slots, yielding provably optimal solutions for coverage and revisit objectives; however, the orbital slots and their underlying geometry (altitude, inclination, eccentricity) must be fixed a priori, so the method cannot discover improved orbit parameters.
We benchmark against SA/GA/DE baselines quantitatively in \secref{sec:baselines}.

\citet{hwang_large-scale_2014} demonstrates true gradient-based optimization of a satellite system through OpenMDAO (Open-Source Framework for Multidisciplinary Design, Analysis, and Optimization), optimizing a single satellite's subsystem design across seven coupled disciplines with over 25{,}000 design variables, with gradient computation made tractable with adjoint state methods. Their work utilizes gradient-based objectives combined with orbit-dependent parameters,  such as solar panel angles, battery sizing, and communication scheduling, to optimize a single spacecraft's operation. 

% This work differs in two respects: we use true analytical gradients through a complete computational graph rather than finite-difference approximations through a simulator, and we make the coverage and revisit objectives themselves differentiable via continuous relaxations rather than treating them as discrete metrics.

\subsection{Differentiable Physics and Simulation}

Differentiable programming~\citep{baydin_automatic_2018} allows for gradients to be obtained across domains even where forward models are complex but smooth. 
The benefits of differentiable programming have been made significantly more accessible through frameworks that perform automatic differentiation, such as PyTorch and JAX.
\citet{hu_difftaichi_2020} demonstrated differentiable physical simulators targeting robotics and fluid dynamics, and \citet{degrave_differentiable_2019} develops differentiable physics engines for contact-rich robotics.
For continuous relaxations applied to parameterized categorical distributions, Gumbel-Softmax reparameterization~\citep{jang_categorical_2017} is frequently used to compute analytic gradients.
\citet{kamra_gradient-based_2021} proves a coverage gradient theorem for spatial multi-resource coverage objectives, providing an estimator for the gradient that can be used for optimization of sensor placement, via spatial discretization and implicit boundary differentiation applied to 2D planar sensor networks.

In astrodynamics, \citet{acciarini_closing_2025} introduces $\partial$SGP4, a PyTorch reimplementation of the SGP4/SDP4 orbit propagation algorithm, which is forward mode differentiable.
\citet{naylor_dlite_2025} extends this direction by combining a JAX-based differentiable SGP4 propagator with Mitsuba~\citep{noauthor_mitsuba_nodate}, a differentiable renderer, to optimize trajectories for spacecraft-to-spacecraft inspection operations, demonstrating end-to-end forward-mode differentiation through orbit propagation chained with a rendering pipeline.

This work builds on $\partial$SGP4 by introducing differentiable coverage and revisit objectives that enable direct optimization of mission performance metrics.

\section{Method}
\label{sec:method}

\subsection{Problem Formulation}
\label{sec:setup}

We consider a constellation of $N$ satellites, each parameterized by a two-line element (TLE) set. 
Each satellite brings up to six degrees of freedom corresponding to their orbital elements, $[a,~e,~i,~\Omega,~\omega,~\mathcal{M}]^\top$, corresponding to semi-major axis, eccentricity, inclination, right ascension of the ascending node (RAAN), argument of periapsis, and mean anomaly, respectively.
We fix the remaining TLE parameters corresponding to mean motion $n$, drag term $B^*$, and secular rates $\dot{n}$, $\ddot{n}$ to keep the constellation configuration static over the optimization horizon (\SI{24}{\hour} in this work).

Given a grid of $J$ ground target points $\mathcal{G} = \{\vect{g}_1, \ldots, \vect{g}_J\}$ on the Earth's surface and a time horizon $[0, T]$ discretized into $K$ steps with spacing $\delta t = T/(K-1)$, we seek to maximize coverage over the Earth and minimize the mean worst-case revisit gap.

\paragraph{Propagation.}
Each satellite's TLE elements are propagated through $\partial$SGP4~\citep{acciarini_closing_2025} to produce position vectors $\vect{r}_i(t_k)$ in the True Equator Mean Equinox (TEME) reference frame.
These are rotated to the Earth-Centered Earth-Fixed (ECEF) frame via the Greenwich Mean Sidereal Time angle $\theta_{\text{GMST}}(t_k)$:
\begin{equation}
    \vect{r}_i^{\text{ECEF}}(t_k) = \mat{R}_z\left(\theta_{\text{GMST}}(t_k)\right) \, \vect{r}_i^{\text{TEME}}(t_k),
    \label{eq:teme_to_ecef}
\end{equation}
where $\mat{R}_z(\theta)$ is the standard rotation matrix about the $z$-axis.
This single-axis rotation is an approximation and neglects polar motion; the complete TEME-to-ECEF transformation requires an additional rotation between the Earth's instantaneous rotation axis and the ECEF pole~\citep{vallado_revisiting_nodate}.
For the \SI{24}{\hour} propagation horizons considered in this work,
polar-motion offsets are on the order of tenths of arcseconds,
corresponding to sub-kilometer position errors. These errors are small
relative to the $\sim$\SI{2200}{\kilo\metre} ground footprint
of a satellite at \SI{550}{\kilo\metre} altitude with
\SI{10}{\degree} minimum elevation. Longer horizons require the full IAU~2006/2000A chain; we return to this in \secref{sec:conclusion}.

\paragraph{Elevation angle.}
The elevation angle from ground point $\vect{g}_j$ to satellite $i$ at time $t_k$ is
\begin{equation}
    \alpha_{ij}(t_k) = \arcsin\left(\frac{(\vect{r}_i^{\text{ECEF}}(t_k) - \vect{g}_j) \cdot \hat{\vect{g}}_j}{\|\vect{r}_i^{\text{ECEF}}(t_k) - \vect{g}_j\|}\right),
    \label{eq:elevation}
\end{equation}
where $\hat{\vect{g}}_j = \vect{g}_j / \|\vect{g}_j\|$ is the local vertical unit vector at the ground point.
A satellite is visible when $\alpha_{ij} \geq \alpha_{\min}$, where we consider the minimum elevation angle $\alpha_{\min} = \SI{10}{\degree}$; the specific threshold depends on the mission type (payload beamwidth, link budget, atmospheric-loss tolerance) and is a user-supplied constant.

\paragraph{Coverage and revisit.}

Instantaneous coverage at ground point $\vect{g}_j$ at time $t_k$ is the indicator function for when at least one satellite is visible: 
\begin{equation}
    C_j(t_k) = \mathbf{1}\!\left[\bigvee_{i=1}^{N} \alpha_{ij}(t_k) \geq \alpha_{\min}\right].
    \label{eq:coverage}
\end{equation}
The instantaneous coverage fraction over the full set of ground points $\mathcal{G}$ is then
\begin{equation}
    C = \frac{1}{KJ} \sum_{k=1}^{K} \sum_{j=1}^{J} C_j(t_k).
    \label{eq:coverage_fraction_hard}
\end{equation}
The revisit gap at ground point $\vect{g}_j$ is the elapsed time since the most recent coverage event:
\begin{equation} 
 \Delta_j(t_k) = t_k - t_{k'}, \quad k' = \max\left\{k'' \leq k : C_j(t_{k''}) = 1\right\}
    \label{eq:revisit}
\end{equation}
and the mean worst-case revisit across all ground points is then
\begin{equation}
    \Delta^{\max} = \frac{1}{J} \sum_{j=1}^{J} \max_k \Delta_j(t_k).
    \label{eq:mean_revisit_hard}
\end{equation}

We now introduce the relaxations required to make these metrics differentiable.

\subsection{Relaxation 1: Soft Visibility}
\label{sec:soft_visibility}

We replace the hard step function with a sigmoid (\figref{fig:relaxations}a):
\begin{equation}
    \tilde{c}_{ij}(t_k) = \sigma\!\left(\frac{\alpha_{ij}(t_k) - \alpha_{\min}}{\tau}\right), \quad \sigma(x) = \frac{1}{1 + e^{-x}},
    \label{eq:soft_visibility}
\end{equation}
where $\tau > 0$ controls the sharpness.
As $\tau \to 0$ this recovers the hard indicator.
This relaxation is analogous to the temperature-scaled softmax in Gumbel-Softmax~\citep{jang_categorical_2017}, with unbounded support and non-zero gradient throughout. This pairs with the noisy-OR of \secref{sec:noisy_or}: the product $\prod_i(1 - \tilde{c}_{ij})$ never zero-clips, and gradients flow through every factor. In contrast, a clamped ramp function saturates to $0$ below a cutoff and loses the gradient for distant satellites. $\tanh$ and $\operatorname{erf}$ also share these properties, but are equivalent to sigmoid up to reparameterization (Appendix~\ref{app:alternatives}).

\subsection{Relaxation 2: Noisy-OR Coverage Aggregation}
\label{sec:noisy_or}

% For a ground point $\vect{g}_j$ at time $t_k$, we need to determine whether \emph{any} satellite provides coverage---a Boolean OR.
Starting with the Boolean OR operator in \eqref{eq:coverage}, treating each $\tilde{c}_{ij}(t_k)$ as an independent detection probability, the noisy-OR~\citep{pearl_probabilistic_2014} relaxation gives (\figref{fig:relaxations}b):
\begin{equation}
    \tilde{C}_j(t_k) = 1 - \prod_{i=1}^{N} \left(1 - \tilde{c}_{ij}(t_k)\right).
    \label{eq:noisy_or}
\end{equation}
This function monotonically increases in each $\tilde{c}_{ij}$, and recovers the hard OR when all inputs are in $\{0, 1\}$.
This relaxation also intrinsically handles overlapping coverage: additional coverage of a point produces diminishing marginal gain, preventing the optimizer from clustering satellites.
Noisy-OR is the unique member of the probabilistic t-conorm family~\citep{klement_triangular_2000} that is smooth, bounded in $[0,1]$, and strictly increasing in each argument; the other smooth alternative, the bounded-sum t-conorm $\min(1, \sum_i \tilde{c}_{ij})$, clips at $1$ and has zero-gradient regions once multiple satellites cover a single point (Appendix~\ref{app:alternatives}).

The soft coverage fraction over the full set of ground points $\mathcal{G}$ with soft visibility and noisy-OR aggregation then becomes
\begin{equation}
    \tilde{C} = \frac{1}{KJ} \sum_{k=1}^{K} \sum_{j=1}^{J} \tilde{C}_j(t_k).
    \label{eq:coverage_fraction}
\end{equation}

\subsection{Relaxation 3: Leaky Integrator for Revisit Gaps}
\label{sec:leaky_integrator}

The revisit gap at a ground point is the time elapsed since the most recent coverage event. Computing this gap requires identifying discrete transitions which we instead approximate with a leaky integrator recurrence~\citep{gerstner_spiking_2002} (\figref{fig:relaxations}c):
\begin{equation}
    \tilde{\Delta}_j(t_k) = \left(\tilde{\Delta}_j(t_{k-1}) + \delta t\right) \cdot \left(1 - \tilde{C}_j(t_k)\right), \quad \tilde{\Delta}_j(t_0) = 0.
    \label{eq:leaky_integrator}
\end{equation}
When the ground point $\mathbf{g}_j$ is covered ($\tilde{C}_j \approx 1$), the accumulated gap is multiplied by approximately zero, resetting it.
When the point is not covered ($\tilde{C}_j \approx 0$), the gap grows by $\delta t$.
This recurrence is fully differentiable and its maximum over time approximates the true worst-case revisit gap at each ground point.
We use this form because the multiplicative reset is exact on binary inputs and keeps the gap in minutes. Alternatives (exponential decay, piecewise-soft reset gates, mean-square-gap aggregates) lose one or the other (Appendix~\ref{app:alternatives}).

\subsection{Relaxation 4: LogSumExp Soft-Maximum}
\label{sec:logsumexp}

\begin{figure}[htb]
    \centering
    % STEP 1: The Images
    % This resizebox forces the whole row of images to be exactly textwidth wide.
    % The "!" tells LaTeX to calculate the height automatically to maintain aspect ratio.
    \resizebox{\textwidth}{!}{%
        \includegraphics[height=3cm]{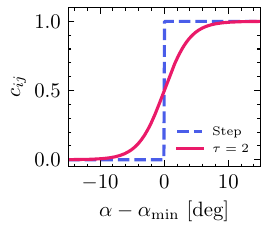}%
        \includegraphics[height=3cm]{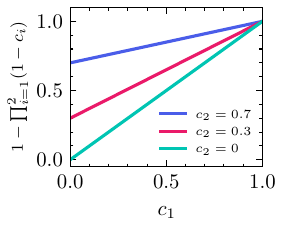}%
        \includegraphics[height=3cm]{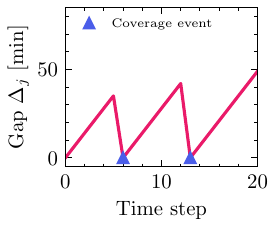}%
        \includegraphics[height=3cm]{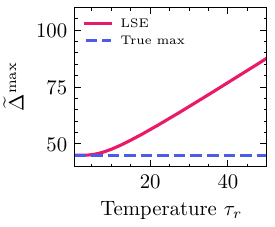}%
    }
    
    % STEP 2: The Captions
    % We use a minipage or makebox to align captions under the images.
    % Since we don't know the exact width of each image above (it was dynamic),
    % we use 0.24\textwidth as a safe approximation for the caption text width.
    \vspace{-2mm} % Pull captions closer if needed
    \begin{center}
        \begin{minipage}{0.24\textwidth}\centering (a) Soft sigmoid \end{minipage}%
        \hfill
        \begin{minipage}{0.24\textwidth}\centering (b) Noisy-OR \end{minipage}%
        \hfill
        \begin{minipage}{0.24\textwidth}\centering (c) Leaky integrator \end{minipage}%
        \hfill
        \begin{minipage}{0.24\textwidth}\centering (d) LogSumExp \end{minipage}%
    \end{center}

    \caption{The four continuous relaxations used in this work to produce a differentiable loss function.}
    \label{fig:relaxations}
\end{figure}

The soft worst-case revisit gap at ground point $\vect{g}_j$ is $\max_k \tilde{\Delta}_j(t_k)$, which has zero gradients almost everywhere.
We replace it with the LogSumExp smooth maximum~\citep{boyd_convex_2023} (\figref{fig:relaxations}d):
\begin{equation}
    \widetilde{\Delta}_j^{\max} = \beta \cdot \log\!\left(\sum_{k=1}^{K} \exp\!\left(\frac{\tilde{\Delta}_j(t_k)}{\beta}\right)\right),
    \label{eq:logsumexp}
\end{equation}
where $\beta$ is a sharpness parameter, for which we assign $\beta = \SI{10}{\minute}$.
This relaxation satisfies the inequality $\max_k \tilde{\Delta}_j(t_k) \leq \widetilde{\Delta}_j^{\max} \leq \max_k \tilde{\Delta}_j(t_k) + \beta \log K$, providing an upper bound.
We use LogSumExp because it is smooth, bounded within $\beta \log K$ of the true maximum, and has a single sharpness parameter $\beta$ in the same units as the gap. Mellowmax differs from LogSumExp by an additive constant and hence admits identical optimizer dynamics. Other soft-max operators ($p$-norm~\citep{boyd_convex_2023}, Boltzmann softmax~\citep{asadi_alternative_2017}) are equally permissible here but are not utilized in this work (Appendix~\ref{app:alternatives}).
The mean worst-case revisit across all ground points is then
\begin{equation}
    \tilde{\Delta}^{\max} = \frac{1}{J} \sum_{j=1}^{J} \widetilde{\Delta}_j^{\max}.
    \label{eq:mean_revisit}
\end{equation}

Coverage and revisit objectives have different sensitivity to the sigmoid temperature $\tau$ in \eqref{eq:soft_visibility}: there exists an additive bias for coverage but a multiplicative bias for the leaky integrator. Hence, it is desirable to separate these parameters. We split $\tau$ into $\tau_{\text{cov}}$ for the coverage objective and $\tau_{\text{rev}}$ for the revisit computation, calibrated in \secref{sec:tightness}.

\subsection{Combined Objective}
\label{sec:objective}

The total loss combines coverage maximization with revisit minimization:
\begin{equation}
    \mathcal{L}(\vect{x}) = -\tilde{C} + \lambda \, \tilde{\Delta}^{\max},
    \label{eq:loss}
\end{equation}
where $\vect{x}$ collects all free and reparameterized (\secref{para:interval_constraints}) orbital parameters and $\lambda$ is a weighting coefficient. The coverage term is instantaneous: it penalizes overlapping fields of view and saturates once satellites are well-separated. The revisit term, aggregated over the full simulation, continues providing gradients after coverage plateaus. Although both terms push toward spacing satellites apart, they operate at different scales---coverage quickly separates clusters, while revisit works globally.

For objectives that are not geodesically uniform, the sums over ground points can be replaced with weighted sums over a discrete target set (\secref{sec:weighted}). 

\subsection{Tightness of the Composed Relaxation}
\label{sec:tightness}

Each of the four relaxations is individually tight: the soft sigmoid recovers the hard step as $\tau \to 0$; the noisy-OR is exact when its inputs are binary; the leaky integrator is exact when coverage is in $\{0, 1\}$; and the LogSumExp satisfies $\max_k x_k \leq \text{LSE} \leq \max_k x_k + \beta \log K$.
However, the composition of these relaxations does not inherit these guarantees, because each relaxation receives the soft output of the previous one rather than the binary values for which it was designed.

The interaction introduces two competing biases:

\paragraph{Phantom coverage (gap shortening).}
The soft sigmoid coverage indicator assigns nonzero coverage $\tilde{c}_{ij} > 0$ to satellites slightly \emph{below} the visibility threshold ($\alpha_{ij} < \alpha_{\min}$).
This phantom coverage propagates through the noisy-OR and partially resets the leaky integrator, producing gaps that are \emph{shorter} than the true discrete gaps.

\paragraph{Incomplete reset (gap lengthening).} \label{para:gap_lengthening}
Conversely, a satellite well above the threshold produces $\tilde{c}_{ij} \approx 0.99$, not exactly 1.
The leaky integrator retains a fraction $(1 - \tilde{C}_j) \approx 0.01$ of the accumulated gap at each coverage event.
Over time, this residual prevents the gap from fully resetting, producing relaxed gaps that are \emph{longer} than the true gaps during sustained coverage.
The LogSumExp adds a further upward bias of at most $\beta \log K$.

These effects compete: phantom coverage shortens gaps near the visibility boundary, while incomplete resets and the LogSumExp lengthen them.
The net bias depends on the specific geometry and is neither a guaranteed upper nor lower bound on the true objective.

The split sigmoid temperatures introduced in \secref{sec:logsumexp} provide a parameter for tuning this effect. Under a shared $\tau$, phantom coverage from clustered below-threshold satellites compounds through the noisy-OR and partially resets the leaky integrator, so the relaxed loss can rank clustered solutions above well-spread ones at the same hard metric. A grid search (Appendix~\ref{app:hyperparam_grid}) shows that $\tau_{\text{cov}} \gg \tau_{\text{rev}}$ restores the correct ordering of converged solutions while keeping smooth coverage gradients. However, this competes with optimization performance: ablation (b) in \secref{sec:ablations} finds that from a reasonable initialization, shared $\tau$ converges better than either split variant we tried. We therefore use shared $\tau$ in all experiments noting that certain problems may benefit from separated temperatures.

\subsection{Optimization}
\label{sec:optimization}

\begin{figure*}[htb]
    \centering
    \includegraphics[width=\linewidth]{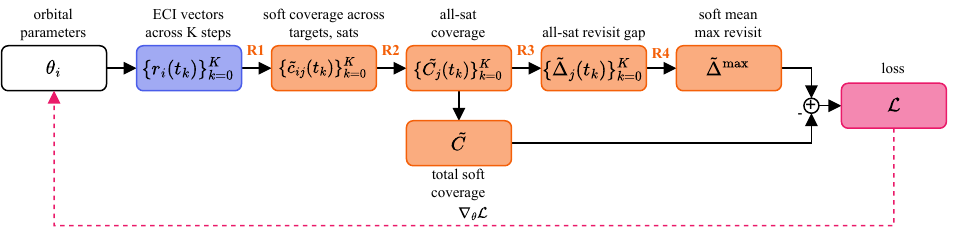}
        \caption{%
        Computational graph for differentiable constellation optimization. Orbital parameters $\boldsymbol{\theta}_i$ are propagated to ECI position vectors $\{\vect{r}_i(t_k)\}$ across $K$ timesteps. Four continuous relaxations (orange, \textbf{R1}--\textbf{R4}) transform the discrete coverage and revisit objectives into relaxed, differentiable objectives:
        \textbf{R1}~soft sigmoid visibility over targets and satellites,
        \textbf{R2}~noisy-OR aggregation to all-satellite coverage,
        \textbf{R3}~leaky integrator revisit gap accumulation, and
        \textbf{R4}~LogSumExp soft-maximum over time.
        The coverage branch produces total soft coverage $\tilde{C}$;
        the revisit branch produces the soft mean worst-case revisit
        $\tilde{\Delta}^{\max}$.
        Both feed into the loss $\mathcal{L}$, and gradients
        $\nabla_{\boldsymbol{\theta}}\mathcal{L}$ flow back via
        reverse-mode automatic differentiation (dashed, pink).%
    }
    \label{fig:graph}
\end{figure*}

\paragraph{Computational graph.}
The full forward pass composes the $\partial$SGP4 propagator with the four relaxations into a single differentiable graph (\figref{fig:graph}).
Every operation is implemented in PyTorch~\citep{paszke_pytorch_2019}, so $\nabla_{\vect{\theta}_i} \mathcal{L}$ is computed via reverse-mode automatic differentiation through the entire graph.
The Jacobian $\partial \vect{r}_i^{\text{TEME}} / \partial \vect{x}_i$ is handled inside $\partial$SGP4: it is dense and time-varying---a small change in RAAN rotates the entire orbital plane, affecting position at all future time steps.

\paragraph{Interval constraints and reparameterization.} \label{para:interval_constraints}

Without constraints, the optimizer trivially raises altitude until the satellite footprint covers the full ground grid. We enforce interval constraints (e.g.\ LEO altitude bounds) via sigmoid reparameterization, mapping unconstrained variables $\vect{\theta}_i$ to valid elements $\vect{x}_i$:
\begin{equation}
    x_d = \ell_d + (u_d - \ell_d) \, \sigma(\theta_d) \quad \Rightarrow \quad \frac{\partial x_d}{\partial \theta_d} = (u_d - \ell_d) \, \sigma(\theta_d)(1 - \sigma(\theta_d)),
    \label{eq:reparam}
\end{equation}
for bounded parameters (e.g.\ inclination) with bounds $[\ell_d, u_d]$, and $x_d = \theta_d$ (identity) for periodic parameters (RAAN, mean anomaly) where wrapping is natural.
This reparameterization eliminates the need for projected gradient or constrained optimization methods.
For coupled constraints, such as requiring a minimum perigee altitude regardless of eccentricity, the per-element sigmoid is insufficient since perigee $r_p = a(1-e)$ couples semi-major axis and eccentricity. For this case, we reparameterize in (perigee altitude, excess altitude) space:
\begin{align}
    r_p &= r_p^{\min} + (r_p^{\max} - r_p^{\min})\,\sigma(\theta_{r_p}), \label{eq:reparam_rp} \\
    \delta r &= \delta r^{\max}\,\sigma(\theta_{\delta r}), \label{eq:reparam_dr} \\
    a &= R_\oplus + r_p + \tfrac{1}{2}\delta r, \quad e = \frac{\delta r}{2a}, \label{eq:reparam_ae}
\end{align}
where $r_p$ is the perigee altitude, $\delta r = r_a - r_p \geq 0$ is the excess altitude (apogee minus perigee), and both are independently sigmoid-bounded. This construction guarantees $r_p \geq r_p^{\min}$ and $e \geq 0$ by construction, while allowing the optimizer to freely explore eccentric orbits.

% \paragraph{Ephemeral graph and gradient bridging.}
% A practical complication is that dSGP4's \texttt{initialize\_tle} function rebuilds the computation graph from scratch at each call, producing a new ephemeral tensor $\vect{x}_i^{\text{eph}}$ for each satellite's elements.
% Standard optimizers like \texttt{torch.optim.Adam} track specific tensor objects and would lose their momentum state across graph rebuilds.
% We solve this by maintaining persistent $\vect{\theta}_i$ tensors that Adam tracks, and bridging gradients from dSGP4's ephemeral tensors after each backward pass via \eqref{eq:reparam}:
% \begin{equation}
%     \vect{\theta}_i.\text{grad} \leftarrow \nabla_{\vect{x}_i^{\text{eph}}} \mathcal{L} \cdot \frac{\partial \vect{x}_i}{\partial \vect{\theta}_i},
%     \label{eq:grad_bridge}
% \end{equation}
% where the Jacobian $\partial \vect{x}_i / \partial \vect{\theta}_i$ is diagonal (\eqref{eq:reparam}).
% Adam then steps on $\vect{\theta}_i$, the updated values are mapped to $\vect{x}_i$ via \eqref{eq:reparam}, and the TLE is re-initialized for the next iteration.

\paragraph{Plane constraints via gradient accumulation.}
Conventionally, satellite constellations are designed with multiple satellites sharing a single plane, to allow for launching multiple satellites per launch vehicle.
Rather than enforcing these per-plane constraints as hard constraints, we average the gradients per-plane: if satellites in a plane share a single RAAN parameter, the gradient is the average of per-satellite contributions.
The averaged gradient is applied to the shared parameter, and the updated value is synchronized across all satellites in the plane after each optimizer step.
This reduces the effective dimensionality without modifying the forward pass, and generalizes to any shared parameters: per-plane inclination sharing, altitude coupling across the constellation, or any grouping governed by mission constraints.

% \paragraph{Computational cost.}
% The per-iteration cost is $\mathcal{O}(NKJ)$: $N$ satellite propagations over $K$ time steps, each evaluated against $J$ ground points.

\section{Experiments}
\label{sec:experiments}

We evaluate the framework on three experiments: a toy problem validating gradient correctness, a uniform-coverage baseline that recovers known-optimal geometries, and a weighted target optimization on a problem class where parametric families fail.

\subsection{Common Setup}
\label{sec:common_setup}

All experiments use $\partial$SGP4~\citep{acciarini_closing_2025} for orbit propagation with PyTorch autograd for gradient computation, and AdamW with learning rate $\eta = 10^{-2}$ for optimization.
Circular orbits are assumed and held stationary by zeroing the mean-motion gradient.
The minimum elevation angle for all ground points is $\alpha_{\min} = \SI{10}{\degree}$.
Coverage and revisit metrics are evaluated over a \SI{24}{\hour} propagation horizon discretized into $K = 240$ time steps on a ground grid of $36 \times 72 = 2592$ points spanning $\pm\SI{70}{\degree}$ latitude, with $\cos(\text{lat})$ weighting on each cell to correct for area distortion.
In each experiment the satellite count $N$ and plane structure are fixed ahead of time and chosen to match the intended comparison geometry (e.g.\ Walker 24/6/1 in \secref{sec:uniform}) as we are primarily looking at configuration and not sizing.
Specific relaxation parameters, optimizer settings, and revisit weight $\lambda$ vary per experiment and are stated in each subsection; \tabref{tab:exp_params} in Appendix~\ref{app:exp_params} collects all per-experiment values in one place.

\subsection{Experiment 1: Toy Problem}
\label{sec:toy}
    
To validate that the relaxed objectives produce correct gradients and results, we consider the simplest non-trivial constellation design problem: two satellites in a single orbital plane at \SI{550}{\kilo\metre} altitude with inclination fixed at \SI{60}{\degree} and RAAN fixed at \SI{0}{\degree}, optimizing only their mean anomalies $\mathcal{M}_1$ and $\mathcal{M}_2$.
We use relaxation parameters $\tau = \SI{2}{\degree}$ (shared), $\beta = \SI{10}{\minute}$, and $\lambda = 2.0$.

The optimal solution is known analytically: the two satellites should be separated by \SI{180}{\degree} in mean anomaly to maximize coverage and revisit rate within the shared orbital plane.
This reduces the problem to two free parameters, allowing the full loss landscape to be visualized.

\begin{figure}[!htbp]
    \centering
    \includegraphics[width=.99\textwidth]{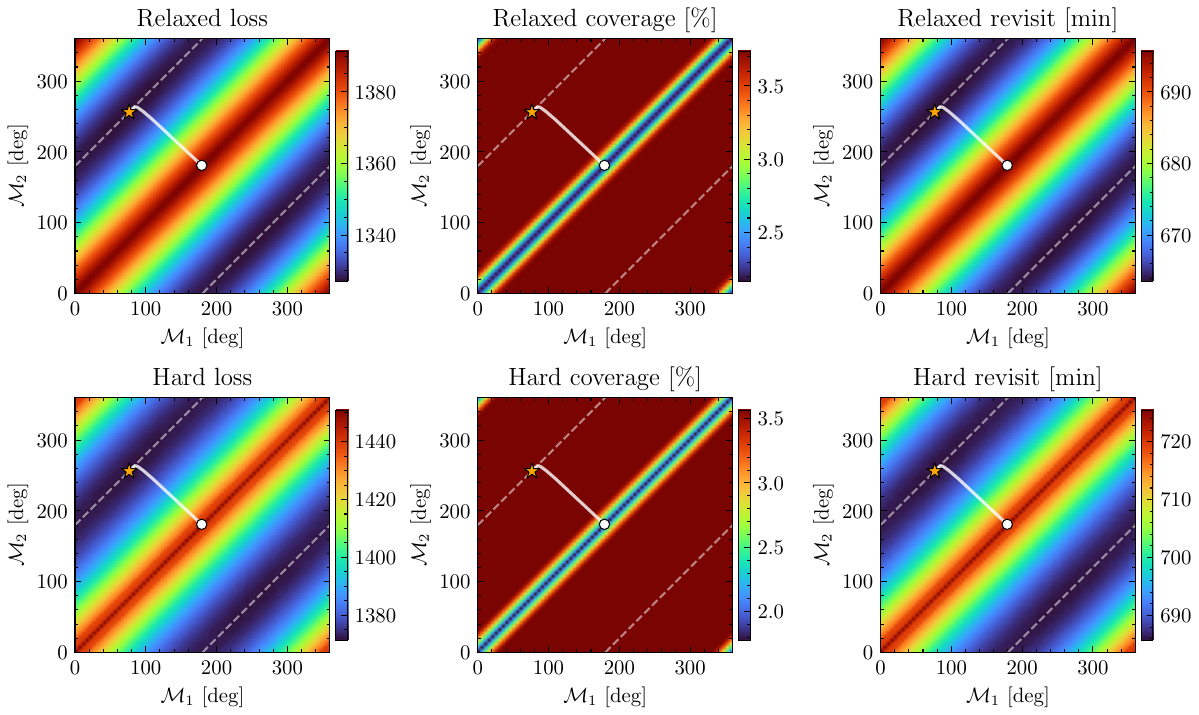}
    \caption{Two-satellite mean anomaly optimization over $(\mathcal{M}_1, \mathcal{M}_2) \in [0, 360)^2$\,\si{\degree}. \textbf{Top row:} relaxed metrics (coverage, revisit, combined loss). \textbf{Bottom row:} hard metrics on the same grid. AdamW optimizer path shown in white, starting from $(\SI{179}{\degree},\SI{181}{\degree})$ with a gold star marking the converged point near the anti-diagonal $|\mathcal{M}_1 - \mathcal{M}_2| = \SI{180}{\degree}$. The relaxed landscape preserves the basin structure of the hard landscape while providing smooth gradients.}
    \label{fig:toy_landscape}
\end{figure}

\paragraph{Loss landscape.}
We evaluate both the relaxed loss and hard (discrete) metrics on a dense grid over $(\mathcal{M}_1, \mathcal{M}_2) \in [0, 360)^2$\SI{}{\degree}.  
\figref{fig:toy_landscape} shows the resulting landscapes.
Both landscapes exhibit a clear global minimum along the anti-diagonal $|\mathcal{M}_1 - \mathcal{M}_2| = \SI{180}{\degree}$, with no local minima.
The relaxed landscape is visibly smoother than the hard landscape while preserving the same basin structure, confirming that the relaxations provide a smooth surface amenable to gradient descent without significantly distorting the underlying objective.

\paragraph{Gradient descent trajectory.}
Starting from a clustered initial configuration where $\mathcal{M}_1 = \SI{179}{\degree}$, $\mathcal{M}_2 = \SI{181}{\degree}$ the optimizer converges to $|\mathcal{M}_1 - \mathcal{M}_2| \approx \SI{180}{\degree}$ within 800 iterations.
The trajectory (overlaid in white on \figref{fig:toy_landscape}) follows the gradient field smoothly toward the minimum without oscillation, validating that the composed relaxations produce well-conditioned gradients.

\subsection{Experiment 2: Uniform Coverage Baseline}
\label{sec:uniform}

\begin{figure}[!htb]
    \centering
    \begin{minipage}[t]{0.48\textwidth}
        \centering
        \includegraphics[width=\textwidth]{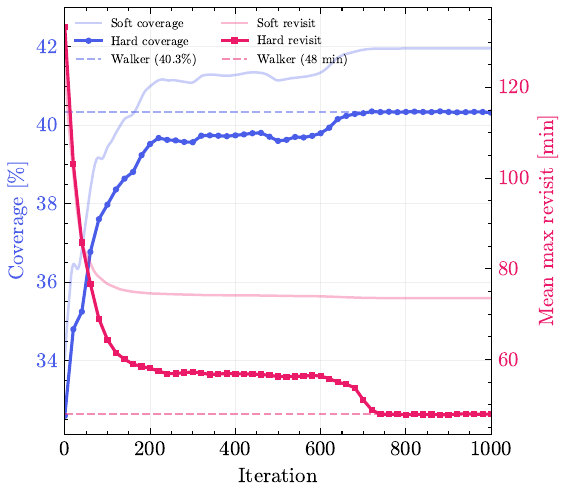}\\[-2pt]
        {\small (a) Convergence}
    \end{minipage}%
    \hfill
    \begin{minipage}[t]{0.48\textwidth}
        \centering
        \includegraphics[width=\textwidth]{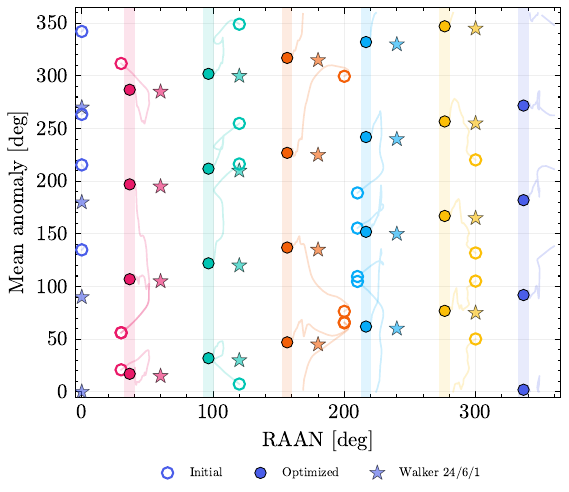}\\[-2pt]
        {\small (b) RAAN vs mean anomaly}
    \end{minipage}
    \caption{Walker recovery experiment. \textbf{(a)} Convergence of hard coverage (left axis) and hard mean worst-case revisit (right axis) over 1000 optimizer iterations, with the Walker 24/6/1 reference values drawn as dashed horizontal lines in the corresponding colors. \textbf{(b)} RAAN vs.\ mean anomaly configuration space: hollow markers are the initial placement of each satellite (irregular RAAN + random MA), filled markers are the optimized placement after 1000 iterations, and stars mark the canonical Walker $24/6/1$ slots. Planes are additionally grouped by color. The optimizer recovers Walker-equivalent geometry up to the rotational symmetry about the $z$-axis.}
    \label{fig:convergence}
\end{figure}

We now apply our gradient-based scheme to a well-understood uniform coverage problem. We optimize $N = 24$ satellites in 6 orbital planes (4 per plane) with inclination fixed at \SI{60}{\degree}, a canonical small-scale Walker configuration with a well-defined optimum (Walker 24/6/1: six RAAN-equispaced planes, 4 satellites per plane, inter-plane phasing $f=1$). 24 satellites is large enough to exhibit the multi-modal basin structure we are interested in while remaining small enough to visualize full RAAN/MA trajectories (\figref{fig:convergence}b) and to serve as a tractable comparison target for the black-box baselines in \secref{sec:baselines}.
RAAN and mean anomaly are free, with RAAN shared within each plane, yielding $6 + 24 = 30$ effective degrees of freedom.

\paragraph{Initial configuration.}
We deliberately initialize the constellation with irregular RAAN offsets $[0, 30, 120, 200, 210, 300]$\SI{}{\degree} far from uniform spacing, and random mean anomalies. The classical Walker-delta $24/6/1$ pattern uses \SI{60}{\degree} RAAN spacing.

\paragraph{Results.}
Over 1000 iterations, the optimizer recovers near-uniform RAAN spacing from the irregular initialization, approaching the Walker-delta geometry without domain-specific inductive biases.
The two well-spread initial planes at \SI{0}{\degree} and \SI{300}{\degree} relax outwards only slightly, while the two clustered pairs at $(30, 120)$\SI{}{\degree} and $(200, 210)$\SI{}{\degree} fan apart to fill the gaps left by the other planes, optimizing to a Walker-Delta \SI{60}{\degree} spacing up to rotational symmetry about the $z$-axis. Metrics are summarized in 
\tabref{tab:uniform_results}, and \figref{fig:convergence} shows convergence, spatial coverage redistribution, and trajectory of each satellite in RAAN/MA configuration space.

\begin{table}[!htb]
\centering
\caption{Recovery of Walker Delta constellation through gradient optimization.}
\label{tab:uniform_results}
\begin{tabular}{lrrr}
\toprule
\textbf{Metric} & \textbf{Initial} & \textbf{Optimized} & \textbf{Walker 24/6/1} \\
\midrule
Hard coverage [\%] & 32.41 & 40.34 & 40.33 \\
Hard mean worst-case revisit [min] & 136.2 & 48.0 & 48.0 \\
\midrule
Soft coverage [\%] & 41.86 & 50.32 & 50.32 \\
Soft mean worst-case revisit [min] & 127.5 & 73.5 & 73.5 \\
\bottomrule
\end{tabular}
\end{table}

\paragraph{Loss landscape.}
To investigate the structure of the loss landscape, we visualize it using per-trajectory PCA~\citep{li_visualizing_2018}: projecting the optimization trajectory for each of the four initializations from the hyperparameter calibration (Appendix~\ref{app:hyperparam_grid}) onto its two principal components and evaluating the loss on that 2D slice.
Within each slice, the relaxed and hard landscapes agree up to smoothing in every case, confirming the relaxations preserve local basin structure. The optimizer descends cleanly: the near-uniform and moderate initializations land in the same low-loss region, while the two clustered starts settle into separate local minima.
The global slice reveals the multi-modal structure of the loss landscape, on a more zoomed out scale than any single PCA slice exposes, with many local minima, and significantly distorted from the hard metric structure.
The smooth local structure means gradient descent is well-conditioned from any start, but initialization determines which local minima the optimizer will descend to, motivating the multi-start ensemble in Ablation~(e) (\secref{sec:ablations}) and framing the Walker recovery of \tabref{tab:uniform_results} as representative of the well-spread-init basin rather than a unilateral guarantee.

\begin{figure*}[!htb]
    \centering
    \includegraphics[width=\textwidth]{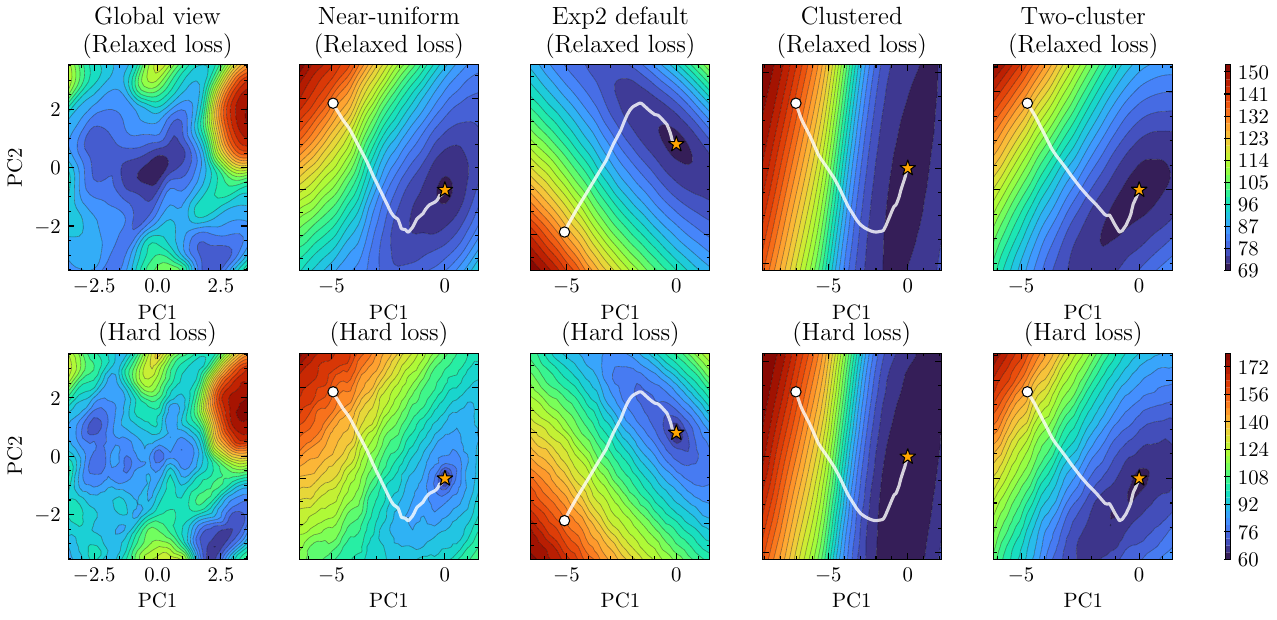}
    \caption{Loss landscape visualization~\citep{li_visualizing_2018}. \textbf{Left column:} global view along two random directions from a Walker-optimal reference point. \textbf{Columns 2--5:} per-trajectory PCA slices for the four calibration initializations (near-uniform, moderate, clustered, two-cluster); each slice projects along that trajectory's two dominant curvature directions, so scales are not comparable across columns. \textbf{Top row:} relaxed loss. \textbf{Bottom row:} hard loss. The color bar beside each panel reports loss value on that 2D projection. The white line is the AdamW optimizer path. Per-trajectory views show smooth basins; the global view reveals multi-modal structure and distinct attractors that account for the init-dependence in Ablation (e).}
    \label{fig:landscape}
\end{figure*}

\subsection{Experiment 3: Regional Target Optimization}
\label{sec:weighted}

An advantage of gradient-based constellation optimization over parametric families is the ability to handle non-uniform, mission-specific objectives.
We demonstrate this by optimizing a small constellation for coverage and revisit over Europe.

\paragraph{Setup.}
We sample 500 ground targets from European continent boundaries, weighted by $\cos(\text{lat})$ to correct for area distortion.
The constellation consists of 4 satellites in 2 orbital planes (2 per plane).
Unlike Experiment~2, we free all orbital geometry parameters: inclination ($\iota \in [30, 90]$\SI{}{\degree}), RAAN, mean anomaly, argument of perigee, and orbital shape via the (perigee, excess altitude) reparameterization of \eqref{eq:reparam_rp}--\eqref{eq:reparam_ae}, with perigee $\in [400, 600]$~\SI{}{\kilo\metre} and excess altitude $\in [0, 1500]$~\SI{}{\kilo\metre}.
Inclination, RAAN, eccentricity, argument of perigee, and altitude are shared within each plane; mean anomaly is free per satellite, giving 12 effective degrees of freedom.
We use $\lambda = 1.0$, and run for 3000 iterations.

\paragraph{Results.}
Starting from a Walker-like circular initialization at \SI{550}{\kilo\metre}, the optimizer discovers a qualitatively different geometry (\tabref{tab:weighted_results}).
Both planes converge to highly elliptical orbits ($e \approx 0.40$) with argument of perigee $\omega \approx \SI{270}{\degree}$, placing apogee over the northern hemisphere where the satellite dwells longest.
The resulting orbits have perigee $\approx$ \SI{685}{\kilo\metre} and apogee $\approx$ \SI{10000}{\kilo\metre}---a Molniya-like highly elliptical orbit discovered entirely from gradients, with no domain-specific inductive biases.
We note that perigee and apogee traverse dramatically different radiation environments---LEO at perigee versus the inner Van Allen belt near apogee---which the current geometry-only loss does not penalize; the sigmoid reparameterization of \secref{para:interval_constraints} admits an apogee cap as an additional constraint (\secref{sec:conclusion}) if desired.

The optimized constellation achieves 99.3\% coverage and 3.7~min mean revisit over Europe, compared to 10.0\% coverage and 247.5~min for a Walker 4/2/1 baseline (\tabref{tab:weighted_results}).
The comparison is not between equal orbit regimes: the Walker baseline is constrained to circular LEO while the optimizer is free to explore eccentric orbits. The result demonstrates that, given this freedom, the optimizer discovers a qualitatively different and more effective geometry for regional coverage.

\begin{figure}[!htb]
    \centering
    \includegraphics[width=.9\textwidth]{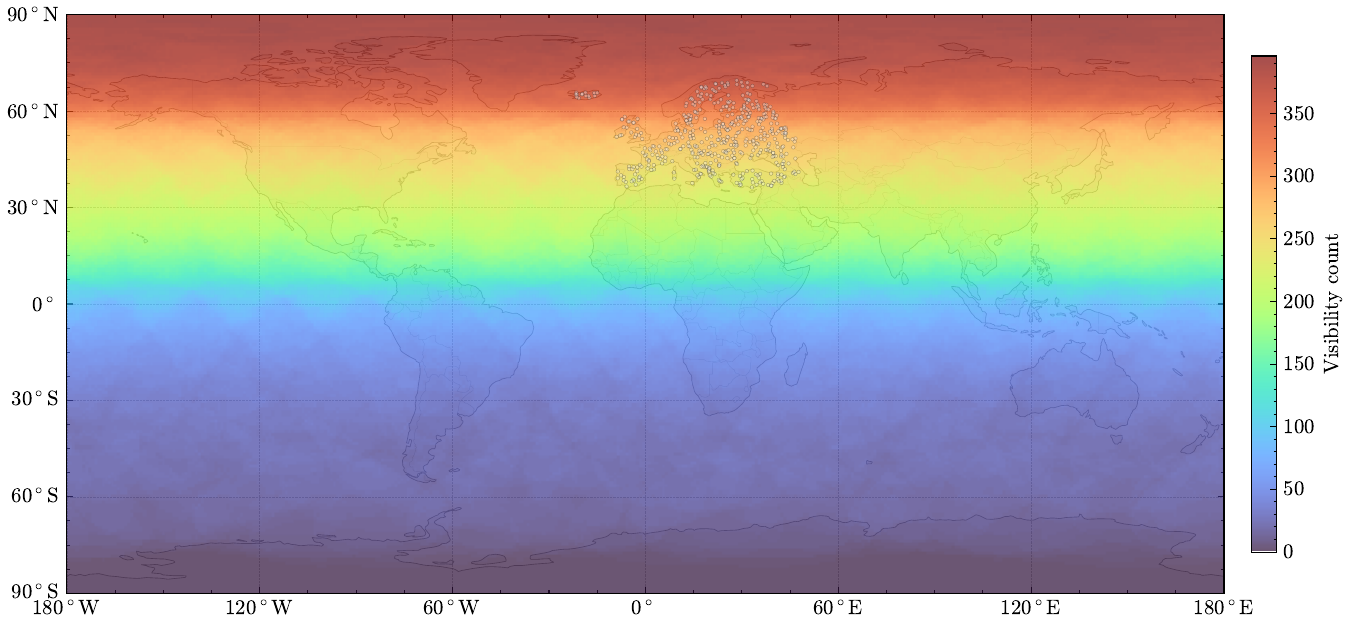}\\[2pt]
    {\small (a) Optimized for Europe}\\[8pt]
    \includegraphics[width=.9\textwidth]{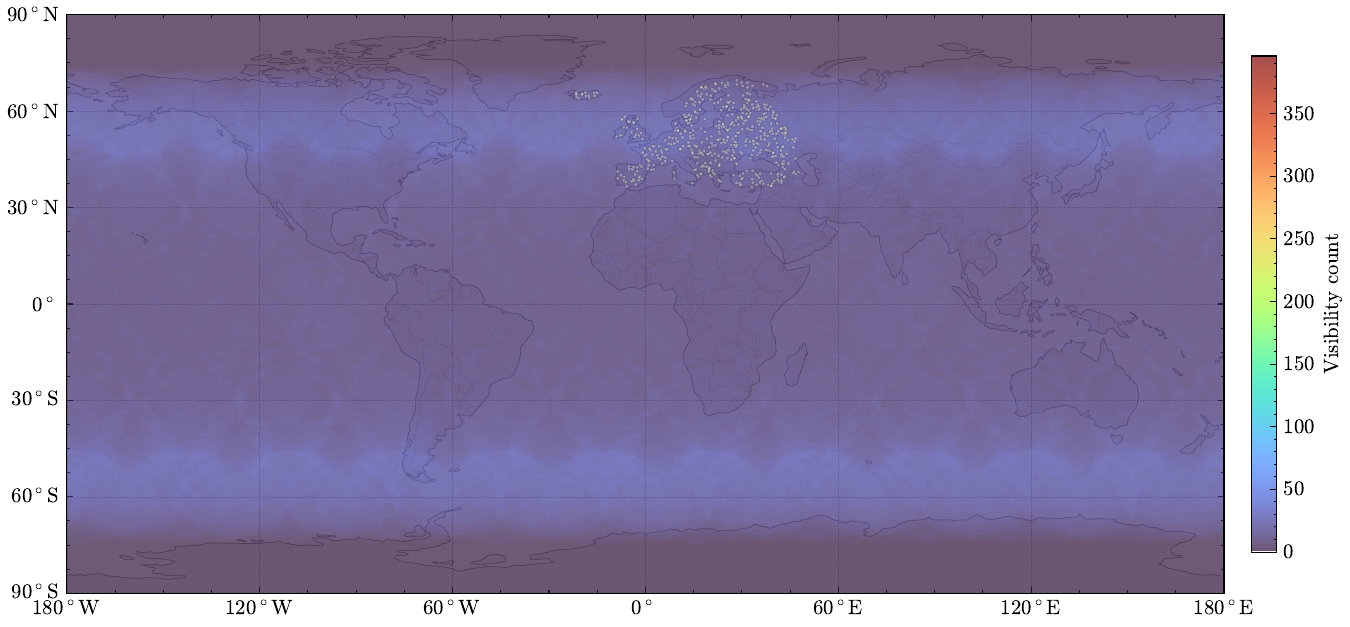}\\[2pt]
    {\small (b) Walker 4/2/1 baseline}
    \caption{Visibility density over \SI{24}{\hour}. \textbf{(a)} The optimized constellation concentrates coverage over Europe via eccentric orbits with apogee dwell over northern latitudes. \textbf{(b)} The Walker baseline distributes coverage uniformly but achieves far less density over the target region.}
    \label{fig:europe_density}
\end{figure}

\figref{fig:europe_density} shows the visibility density (number of timesteps each ground bin is visible to any satellite).
The optimized constellation concentrates nearly all coverage in the northern hemisphere over Europe, while the Walker baseline distributes coverage uniformly but achieves far less density.
The orbital period of $\sim$\SI{210}{\minute} produces approximately 7 revolutions per sidereal day, creating a near-resonant repeating ground track.

\begin{table}[!htb]
\centering
\caption{Regional target optimization: 4 satellites (2 planes $\times$ 2), 3000 iterations, all orbital geometry parameters (aside from fixed planes) free with coupled perigee/apogee constraints.}
\label{tab:weighted_results}
\begin{tabular}{lrr}
\toprule
\textbf{Metric} & \textbf{Walker 4/2/1} & \textbf{Optimized} \\
\midrule
Hard coverage [\%] & 9.97 & 99.25 \\
Hard mean worst-case revisit [min] & 247.5 & 3.7 \\
Soft coverage [\%] & 12.37 & 97.99 \\
Soft mean worst-case revisit [min] & 208.9 & 54.9 \\
\bottomrule
\end{tabular}
\end{table}

\begin{table}[!htb]
\centering
\caption{Per-plane orbital geometry discovered by the optimizer for the Europe target set. Both planes converge to eccentric, Molniya-like orbits with apogee over the northern hemisphere ($\omega \approx \SI{270}{\degree}$). The Walker baseline uses circular orbits at \SI{550}{\kilo\metre}.}
\label{tab:exp3_geometry}
\begin{tabular}{lrrrrr}
\toprule
& $\iota$ \textbf{[deg]} & $\Omega$ \textbf{[deg]} & $e$ & $\omega$ \textbf{[deg]} & \textbf{Perigee / Apogee [km]} \\
\midrule
Plane 0 & 62.6 & 336.6 & 0.398 & 276.2 & 685 / 10011 \\
Plane 1 & 65.3 & 155.6 & 0.402 & 266.5 & 687 / 10193 \\
\midrule
Walker (all) & 60.0 & uniform & 0.001 & --- & 550 / 550 \\
\bottomrule
\end{tabular}
\end{table}

\begin{figure}[htbp]
    \centering
    \includegraphics[width=0.48\textwidth]{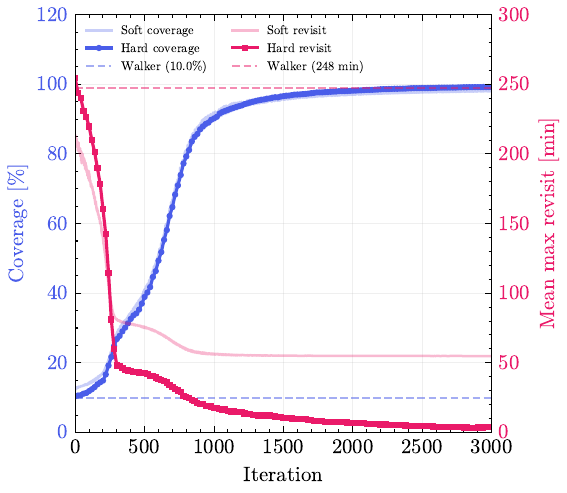}
    \caption{Convergence of the Europe target optimization over 3000 iterations. The optimizer rapidly improves both coverage and revisit over the Walker baseline (dashed).}
    \label{fig:exp3_convergence}
\end{figure}

\subsection{Comparison to Baseline Methods}
\label{sec:baselines}

Constellation configuration in prior work is typically posed as a black-box optimization over design variables, handled either by population heuristics (simulated annealing, genetic algorithms) or gradient-free optimization.
We benchmark our differentiable pipeline against three such baselines on the Walker recovery experiment of \secref{sec:uniform} as it admits a known optimal solution.

\paragraph{Baselines.}
We compare against simulated annealing (SA), a genetic algorithm (GA), and differential evolution (DE), all applied to the hard fitness $\mathcal{L} = -C + \lambda \Delta^{\max}$ computed from \eqref{eq:coverage_fraction_hard} and \eqref{eq:mean_revisit_hard}; the baselines do not use the relaxed objective at all.
SA and GA follow the constellation-design formulations used in prior heuristic work~\citep{paek_optimization_2019} with standard tuning: SA uses a probe-calibrated temperature schedule (\SI{80}{\percent} initial / \SI{1}{\percent} final acceptance, geometric cooling~\citep{kirkpatrick_optimization_1983}) and an adaptive step size targeting the Vanderbilt--Louie rate of $0.44$~\citep{vanderbilt_monte_1984}; GA uses tournament selection~\citep{miller_genetic_nodate} (size 3), uniform crossover~\citep{syswerda_uniform_1989}, and Gaussian mutation with $\sigma$ annealed from $\SI{30}{\degree}$ to $\SI{5}{\degree}$~\citep{back_evolutionary_1996}.
DE uses the TwoPointsDE implementation from the nevergrad library~\citep{rapin_nevergrad_2019}.
All three warm-start from the same initialization as the gradient run and each is run for 5 seeds with a budget of 4050 evaluations (4000 main + the 50-evaluation SA probe, matched across methods), picking the best out of each run.

\begin{table}[!htb]
\centering
\caption{Final hard metrics on the Walker recovery testbed. The gradient pipeline recovers the Walker reference exactly within 1000 iterations; the three black-box baselines plateau well short of it even with roughly four times the evaluation budget and best-of-5 seed selection.}
\label{tab:baselines}
% Manually generated from experiments/exp2_baselines.py (5 seeds, 4050 evals each: 4000 main + 50-eval SA probe, matched across methods)
% Range is min--max across seeds; the best side of each range is bolded
% (highest coverage, lowest revisit).
\begin{tabular}{lrr}
\toprule
\textbf{Method} & \textbf{Hard coverage [\%]} & \textbf{Hard mean worst-case revisit [min]} \\
\midrule
Walker 24/6/1 reference & 40.33 & 48.0 \\
Gradient (ours, 1000 iters) & \textbf{40.34} & \textbf{48.0} \\
\midrule
Simulated annealing       & 34.33--\textbf{36.54} & \textbf{73.9}--80.5 \\
Genetic algorithm          & 37.83--\textbf{38.69} & \textbf{60.3}--62.2 \\
TwoPointsDE~\citep{rapin_nevergrad_2019} & 35.77--\textbf{37.44} & \textbf{68.5}--76.1 \\
\bottomrule
\end{tabular}

\end{table}

\begin{figure}[!htb]
    \centering
    \begin{minipage}[t]{0.49\linewidth}\centering
        \includegraphics[width=\textwidth]{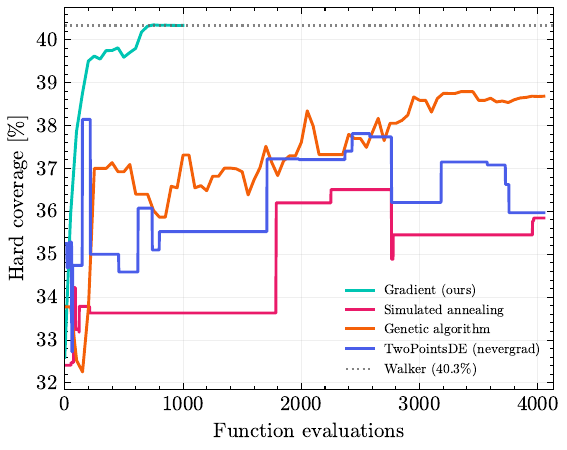}\\[-2pt]
        {\small (a) Hard coverage}
    \end{minipage}\hfill
    \begin{minipage}[t]{0.49\linewidth}\centering
        \includegraphics[width=\textwidth]{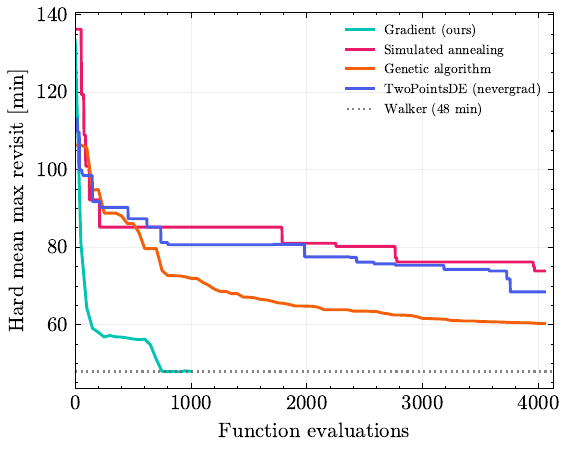}\\[-2pt]
        {\small (b) Hard mean worst-case revisit}
    \end{minipage}
    \caption{Convergence of the differentiable optimizer (teal, Exp 2 baseline) against three black-box baselines on the Walker recovery testbed. Best-of-5 seeds shown per method; evaluation $0$ shares the common irregular-RAAN initialization by zero-order hold, and the Walker reference is dotted grey. The gradient pipeline matches Walker at ${\sim}750$ evaluations; the heuristics plateau above Walker revisit even with roughly four times the budget.}
    \label{fig:baselines_conv}
\end{figure}

\figref{fig:baselines_conv} plots convergence for the best-fitness seed of each method against function evaluations. The gradient-based method reaches Walker-Delta metrics within ${\sim}750$ evaluations; GA plateaus at \SI{60.3}{\minute} revisit, DE at \SI{68.5}{\minute}, and SA at \SI{73.9}{\minute}, none getting close to Walker-Delta even at four times the computational budget.
GA's edge over DE on this problem is attributable to its uniform crossover and $\bmod\,\SI{360}{\degree}$ wrap, which respect the cyclic nature of RAAN and mean anomaly; DE's Euclidean-space perturbations do not.

\newpage
\subsection{Ablations}
\label{sec:ablations}

To isolate the design choices behind the relaxed pipeline, we re-run the Walker recovery testbed of \secref{sec:uniform} under five controlled perturbations. Each row of \tabref{tab:ablations} reports the final hard coverage and mean worst-case revisit; the Walker 24/6/1 reference and results from \secref{sec:uniform} are included for calibration.

\begin{table}[!htb]
\centering
\caption{Design-choice ablations on the Walker recovery testbed (\secref{sec:uniform}, 24 satellites, 1000 iterations, hard metrics evaluated at the final step). Default settings (used elsewhere unless ablated) are $\lambda = 0.1$, $\tau_{\text{cov}} = \tau_{\text{rev}} = \SI{2}{\degree}$, $\beta = \SI{10}{\minute}$.}
\label{tab:ablations}
% Auto-generated by experiments/exp4_ablations.py
\begin{tabular}{lrr}
\toprule
\textbf{Setting} & \textbf{Hard coverage [\%]} & \textbf{Hard mean worst-case revisit [min]} \\
\midrule
Walker 24/6/1 reference / Exp 2 defaults & 40.34 & 48.0 \\
\midrule
(a) Coverage only ($\lambda = 0$) & 40.76 & 126.7 \\
(a) Revisit only ($-\tilde{C}$ dropped) & 40.34 & 48.0 \\
(b) Split $\tau$ ($\tau_{\text{cov}}=3^{\circ}$, $\tau_{\text{rev}}=1^{\circ}$) & 39.04 & 54.9 \\
(b) Split $\tau$ inverted ($\tau_{\text{cov}}=1^{\circ}$, $\tau_{\text{rev}}=3^{\circ}$) & 40.11 & 58.3 \\
(c) $\lambda = 0.01$ & 40.06 & 59.8 \\
(c) $\lambda = 1.0$ & 40.34 & 47.9 \\
(d) $\beta = 1$ min & 38.58 & 55.3 \\
(d) $\beta = 100$ min & 40.08 & 68.9 \\
(e) Random RAAN+MA init, $n=10$ & 39.73 $\pm$ 0.46 (38.82--40.34) & 61.7 $\pm$ 4.7 (47.8--64.1) \\
\bottomrule
\end{tabular}

\end{table}

\paragraph{(a) Loss composition.}
Coverage-only surpasses Walker-level coverage (\SI{40.76}{\percent} vs \SI{40.34}{\percent}) but leaves revisit at \SI{126.7}{\minute}---over 2.6 times the Walker baseline---since the loss never penalizes long gaps.
In contrast, revisit-only and the full combined loss both converge to the Walker-equivalent solution (\SI{40.34}{\percent} coverage, \SI{48.0}{\minute} revisit): once satellites are spaced to minimize revisit, the coverage term contributes only a small residual gradient, so the two are effectively equivalent on this testbed.

\paragraph{(b) Sigmoid temperature.}
We compare the default $\tau = \SI{2}{\degree}$ (shared between coverage and revisit branches) against two variants with decoupled coverage and revisit temperatures: ($\tau_{\text{cov}} = \SI{3}{\degree}$, $\tau_{\text{rev}} = \SI{1}{\degree}$) and the inverse ($\tau_{\text{cov}} = \SI{1}{\degree}$, $\tau_{\text{rev}} = \SI{3}{\degree}$). The shared setting recovers Walker (\SI{40.3}{\percent}, \SI{48.0}{\minute}); the two decoupled variants underperform (\SI{39.0}{\percent} / \SI{54.9}{\minute} and \SI{40.1}{\percent} / \SI{58.3}{\minute}). Shared $\tau$ is the default for all other experiments.

\paragraph{(c) Revisit weight $\lambda$.}
Sweeping $\lambda \in \{0.01, 0.1, 1.0\}$ shows that results are insensitive to $\lambda$ over two orders of magnitude: both $\lambda = 0.1$ and $\lambda = 1.0$ recover Walker (\SI{40.3}{\percent}, \SI{48.0}{\minute}).
Under-weighting revisit at $\lambda = 0.01$ lets coverage saturate before the revisit gradient does useful work (\SI{40.1}{\percent}, \SI{59.8}{\minute}). The ablation suggests a region of applicability rather than a sharp optimum, as similarly evidenced by ablation (a).

\paragraph{(d) LogSumExp temperature $\beta$.}
Sweeping $\beta \in \{1, 10, 100\}$ minutes reveals both ends of the smoothness/correctness trade-off.
The tight $\beta = \SI{1}{\minute}$ approaches a hard max and starves near-worst timesteps of gradient, degrading coverage to \SI{38.6}{\percent}.
The loose $\beta = \SI{100}{\minute}$ over-smooths the worst-case signal so the optimizer underweights long gaps, raising revisit to \SI{68.9}{\minute}.
$\beta = \SI{10}{\minute}$ sits between these failure modes and recovers Walker exactly. From this ablation, it can be observed $\beta$ is likely the most sensitive hyperparameter.

\paragraph{(e) Initialization sensitivity.}
Re-running the same configuration under $n = 10$ seeds with both RAAN and mean anomaly drawn uniformly at random yields hard coverage in $[38.8, 40.3]\,\%$ and revisit in $[47.8, 64.1]$~min, with mean $39.7 \pm 0.5\,\%$ and $61.7 \pm 4.7$~min.
Only one of the ten seeds recovers Walker exactly; the other nine converge to a worse local minimum near \SI{63}{\minute} revisit while still reaching roughly Walker-level coverage.
This finding is consistent with the multi-modal structure visible in \figref{fig:landscape}: coverage drives satellites apart to near-uniform angular spacing reliably, but the finer-grained plane-phasing that distinguishes the global Walker minimum from nearby local minima is not always found from arbitrary RAAN+MA initializations.
The result of \secref{sec:uniform} is therefore not guaranteed from every initialization, but remains robustly reachable with correct choice of initialization, and can be mitigated with parallel runs.

\begin{figure}[!htb]
    \centering
    \begin{minipage}[t]{0.32\textwidth}\centering
        \includegraphics[width=\textwidth]{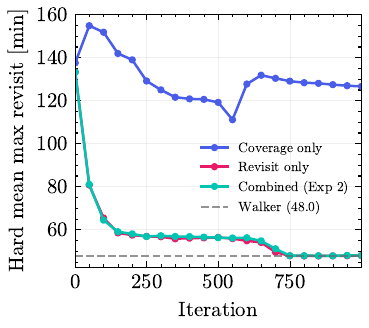}\\[-2pt]
        {\small (a) Loss composition: revisit}
    \end{minipage}\hfill
    \begin{minipage}[t]{0.32\textwidth}\centering
        \includegraphics[width=\textwidth]{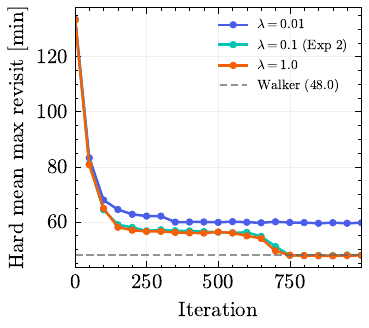}\\[-2pt]
        {\small (b) Revisit weight $\lambda$: revisit}
    \end{minipage}\hfill
    \begin{minipage}[t]{0.32\textwidth}\centering
        \includegraphics[width=\textwidth]{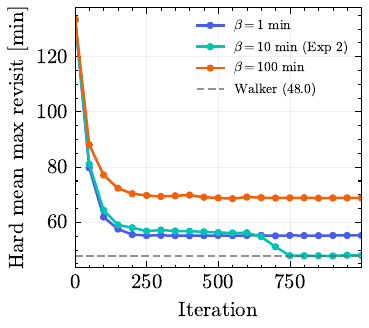}\\[-2pt]
        {\small (c) LSE temperature $\beta$: revisit}
    \end{minipage}
    \\[6pt]
    \begin{minipage}[t]{0.32\textwidth}\centering
        \includegraphics[width=\textwidth]{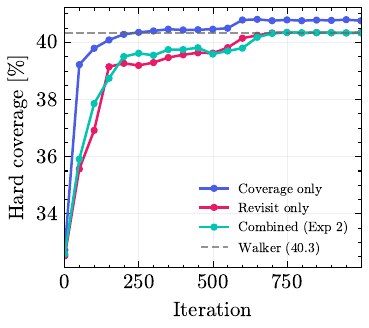}\\[-2pt]
        {\small (d) Loss composition: coverage}
    \end{minipage}\hfill
    \begin{minipage}[t]{0.32\textwidth}\centering
        \includegraphics[width=\textwidth]{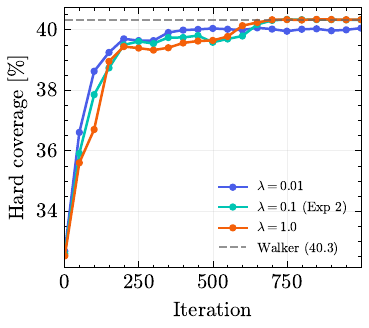}\\[-2pt]
        {\small (e) Revisit weight $\lambda$: coverage}
    \end{minipage}\hfill
    \begin{minipage}[t]{0.32\textwidth}\centering
        \includegraphics[width=\textwidth]{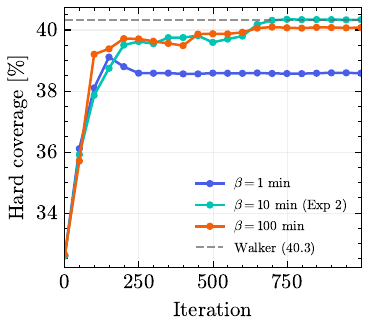}\\[-2pt]
        {\small (f) LSE temperature $\beta$: coverage}
    \end{minipage}
    \caption{Hard metrics over training for the three multi-variant ablations; Walker 24/6/1 reference dashed. \textbf{Top row:} mean worst-case revisit. \textbf{Bottom row:} coverage. \textbf{(a, d)} Coverage-only flattens coverage and surpasses Walker, but never reduces revisit; revisit-only and combined both converge to the Walker solution. \textbf{(b, e)} $\lambda = 0.1$ and $\lambda = 1.0$ are indistinguishable on this testbed; $\lambda = 0.01$ plateaus early on revisit. \textbf{(c, f)} $\beta = \SI{10}{\minute}$ converges to Walker solution; $\beta = \SI{1}{\minute}$ is gradient-starved on revisit and leaves coverage short; $\beta = \SI{100}{\minute}$ over-smooths and gets stuck at a worse revisit minimum.}
    \label{fig:ablations_conv}
\end{figure}

\subsection{Computational Cost and Scalability}
\label{sec:cost}

We report computational cost in terms of evaluations (forward passes) and iterations (forward + backward passes). The per-step cost of the differentiable pipeline scales as $\mathcal{O}(NKJ)$---linear in satellite count $N$, time steps $K$, and ground targets $J$---with the backward pass a small constant multiple of the forward pass.
A single forward pass of our pipeline is a constant factor cheaper than one STK orbit simulation used in prior work~\citep{paek_optimization_2019}, since $\partial$SGP4 is a pure-PyTorch reimplementation of SGP4 rather than a full high-fidelity simulator.

\paragraph{Information per step.}
Each gradient step returns an $N_{\text{dof}}$-dimensional descent direction via autodiff (30-dimensional on Exp 2, 12-dimensional on Exp 3), whereas each black-box evaluation returns a single scalar.
At a per-step cost roughly an order of magnitude larger than a forward-only evaluation, the gradient pipeline delivers $N_{\text{dof}}$ signals per evaluation-equivalent. This is the regime in which gradient methods beat dimension-agnostic heuristics: empirically, the pipeline reaches the Walker reference within ${\sim}750$ iterations (\figref{fig:baselines_conv}), while SA/GA/DE plateau \SI{12}{\minute}--\SI{26}{\minute} worse on revisit even at ${\sim}4\times$ the evaluation budget (\tabref{tab:baselines}).

\paragraph{Scaling to larger problems.}
The problem sizes in this paper fit comfortably on a single GPU. Much larger constellations (hundreds to thousands of satellites) would admit satellite-wise sharding across devices with data-parallel strategies, and multi-start ensembles across initializations parallelize trivially across the independent-seed dimension.

\section{Conclusions and Future Work}
\label{sec:conclusion}

We show that gradient-based optimization is viable for satellite constellation configuration when coverage and revisit metrics are made differentiable through continuous relaxations and composed with a differentiable orbit propagator.
The relaxed objectives are locally tight with their hard counterparts (\secref{sec:tightness}) and produce well-conditioned gradients where finite-difference approaches do not~\citep{paek_optimization_2019}.
On the Walker recovery testbed (\secref{sec:uniform}), the method recovers the Walker 24/6/1 constellation geometry (within rotational symmetry) within ${\sim}750$ iterations, whereas tuned simulated annealing, genetic algorithm, and differential evolution baselines plateau at \SI{60}{\minute}--\SI{74}{\minute} revisit despite ${\sim}4\times$ the evaluation budget (\secref{sec:baselines}, \tabref{tab:baselines}).
On the regional coverage problem (\secref{sec:weighted}), the same pipeline discovers Molniya-like highly elliptical orbits for coverage at extreme latitudes, without any domain-specific inductive biases.

\paragraph{Limitations.}
This work addresses configuration only---satellite count $N$ and plane structure are fixed inputs, with sizing handled outside the gradient loop by search or similar methods. The loss is geometry-only: visibility gates on satellite elevation alone and treats coverage as single-pass, so Sun-illumination for optical payloads and $K$-fold multi-view requirements are not captured. The \SI{24}{\hour} propagation horizon misses long-term perturbations ($J_2$ RAAN drift, atmospheric drag, seasonal repeat tracks), and the single-axis TEME-to-ECEF rotation of \eqref{eq:teme_to_ecef} accumulates kilometer-level ground error beyond ${\sim}$\SI{7}{days} as polar motion grows. Exp 3 also shows that the unconstrained optimum can leave the operational regime: the Molniya-like orbits it discovers cross the inner Van Allen belt near \SI{10000}{\kilo\metre} apogee, which the geometry-only loss does not penalize, but other mission-level requirements such as radiation tolerance may force the excess altitude lower. Of these, Sun-illumination, multi-pass coverage, and operational-regime bounds are simple extensions to the current method (detailed below); sizing, the full IAU~2006/2000A rotation chain, and $J_2$-aware long-horizon propagation require additional investigation.

\paragraph{Future work.}
Several natural directions extend the pipeline without architectural change. The sigmoid reparameterization of \secref{para:interval_constraints} admits any box constraint on the orbital elements, so adding an apogee cap below the inner-belt threshold (an interval constraint on $r_a = a(1+e)$, composable with \eqref{eq:reparam_ae}) or restricting inclination to a launch-available band (an interval constraint on $i$) is immediate; a softer launch-feasibility signal would be a differentiable $\Delta v$ penalty from a nominal dropoff orbit (e.g.\ \SI{500}{\kilo\metre} circular at the launch inclination) to the optimized elements. For mission-level visibility, Sun-elevation above the target~\citep{vallado_fundamentals_2022} would enter as another soft sigmoid composed into the noisy-OR, making daylight-only coverage a product of two soft visibilities, and $K$-fold multi-pass requirements would aggregate soft visibilities with a second LogSumExp (or a $p$-norm soft-max when conditioning gets tight). For longer horizons, parallel multi-epoch aggregation or incorporating $J_2$ secular rates into the propagator would let the optimizer exploit RAAN drift and Sun-synchronous repeats directly; beyond the weeks-or-longer scale, the full IAU~2006/2000A precession--nutation chain would replace the single-axis rotation of \eqref{eq:teme_to_ecef}. Finally, the pipeline is objective-agnostic: heterogeneous sensor placement, constellation reconfiguration, and multi-objective coverage with mission-specific elevation constraints are all expressible as weighted sums over the same graph, but are outside the scope of this work and require further investigation.

\FloatBarrier
\bibliographystyle{unsrtnat}
\bibliography{relaxed_eo}

\appendix

\section{Per-Experiment Parameters}
\label{app:exp_params}

\tabref{tab:exp_params} collects the per-experiment values for reproducibility. Values shared across experiments---propagation horizon \SI{24}{\hour}, $K=240$ time steps, minimum elevation $\alpha_{\min} = \SI{10}{\degree}$, AdamW with learning rate $\eta=10^{-2}$---are omitted. The ground grid is $36 \times 72$ except where a regional target set is specified.

\begin{table}[!htb]
\centering
\caption{Per-experiment parameters for the experiments in \secref{sec:experiments}.}
\label{tab:exp_params}
\begin{tabular}{lccccccc}
\toprule
\textbf{Experiment} & $N$ & \textbf{DoF} & $J$ & $\tau_{\text{cov}}$ [\si{\degree}] & $\tau_{\text{rev}}$ [\si{\degree}] & $\beta$ [\si{\minute}] & $\lambda$ \\
\midrule
Exp 1 (toy)           &  2 &  2 & 2592 & 2.0 & 2.0 & 10 & 2.0 \\
Exp 2 (Walker)        & 24 & 30 & 2592 & 2.0 & 2.0 & 10 & 0.1 \\
Exp 3 (Europe)        &  4 & 12 &  500 & 2.0 & 2.0 & 10 & 1.0 \\
Baselines (SA/GA/DE)  & 24 & 30 & 2592 & --- & --- & ---& 0.1 \\
Hyperparam tuning     & 24 & 30 & 2592 & 2.0 & 2.0 & 10 & 0.1 \\
\bottomrule
\end{tabular}
\end{table}

\section{Hyperparameter Tuning}
\label{app:hyperparam_grid}

This appendix provides the complete experimental setup and raw results for hyperparameter tuning described in \secref{sec:tightness}.

\paragraph{Setup.}
The tuning uses the same 24-satellite setup as \secref{sec:uniform} (see \tabref{tab:exp_params}), run for 800 iterations per initial configuration with shared $\tau = \SI{2}{\degree}$ and GMST randomization enabled. These runs use the \emph{default} relaxation parameters, not the calibrated ones---the goal is to obtain four diverse final solutions whose relative quality is known from the hard metrics.

\paragraph{Initial RAAN configurations.}
The four configurations are: near-uniform $[0, 60, 120, 180, 240, 300]$\SI{}{\degree}, moderate $[0, 30, 120, 200, 210, 300]$\SI{}{\degree}, clustered $[0, 10, 20, 30, 40, 50]$\SI{}{\degree}, and two-cluster $[0, 5, 180, 185, 270, 275]$\SI{}{\degree}.
All use the same random mean anomaly initialization (NumPy \texttt{RandomState(42)}, uniform on $[0, 360)$\SI{}{\degree}).

\paragraph{Grid search parameters.}
The grid search evaluates the relaxed loss of all four final solutions under each combination of: coverage softness $\tau_{\text{cov}} \in \{1.0, 1.5, 2.0, 2.5, 3.0, 4.0, 5.0\}$\SI{}{\degree}, revisit softness $\tau_{\text{rev}} \in \{0.1, 0.25, 0.5, 0.75, 1.0, 1.5, 2.0\}$\SI{}{\degree}, LogSumExp temperature $\beta \in \{3.0, 5.0, 7.5, 10.0, 15.0, 20.0\}$\SI{}{\minute}, and revisit weight $\lambda \in \{0.05, 0.1, 0.2, 0.5, 1.0, 2.0\}$.
This gives $7 \times 7 \times 6 \times 6 = 1764$ combinations, each requiring 4 gradient-free forward passes (one per solution), for 7056 evaluations total.

\paragraph{Validity criterion.}
Each of the four optimizations converges to a different local minimum. A hyperparameter combination is \emph{valid} if the relaxed loss at these local minima preserves the same ordering as the hard loss---i.e., solutions that are better under the hard metrics should also be better under the relaxed loss. Since the hard metrics group the four solutions into two tiers (near-uniform and moderate are better; clustered and two-cluster are worse), validity requires:
\begin{equation}
    \max\bigl(\tilde{\mathcal{L}}_{\text{near-uniform}},\, \tilde{\mathcal{L}}_{\text{moderate}}\bigr) < \min\bigl(\tilde{\mathcal{L}}_{\text{clustered}},\, \tilde{\mathcal{L}}_{\text{two-cluster}}\bigr).
    \label{eq:validity}
\end{equation}
The \emph{margin} is the difference between the right- and left-hand sides.
Among valid combinations, we utilize the one that maximizes $\tau_{\text{cov}}$ (smoothest coverage gradients), then $\tau_{\text{rev}}$, with margin as tiebreaker.

\paragraph{Reproducibility.}
All experiments can be reproduced from the repository found at \url{https://github.com/shreeyam/differentiable_eo}.

\section{Relaxation Equivalences}
\label{app:alternatives}

Several natural alternatives to R1, R2, and R4 reduce to the chosen relaxations up to reparameterization or additive constants. Hence, swapping in the alternatives would not produce different optimizer dynamics.

\subsection{R1: tanh $\Leftrightarrow$ sigmoid}

For any $x \in \mathbb{R}$,
\begin{equation}
    \tfrac{1}{2}\bigl(1 + \tanh(x/2)\bigr)
    = \tfrac{1}{2}\left(1 + \frac{e^{x/2} - e^{-x/2}}{e^{x/2} + e^{-x/2}}\right)
    = \frac{e^{x/2}}{e^{x/2} + e^{-x/2}}
    = \frac{1}{1 + e^{-x}}
    = \sigma(x),
    \label{eq:tanh_sigmoid}
\end{equation}
so the rescaled tanh $\bigl(1 + \tanh((\alpha_{ij} - \alpha_{\min})/(2\tau))\bigr)/2$ is pointwise identical to the sigmoid in \eqref{eq:soft_visibility}.
Gradients, level sets, and every downstream quantity are therefore identical.

\subsection{R2: Noisy-OR $\Leftrightarrow$ probabilistic t-conorm}

The probabilistic t-conorm~\citep{klement_triangular_2000} of $n$ values $\{x_i\} \subset [0,1]$ is defined as $1 - \prod_i (1 - x_i)$, which is exactly the noisy-OR of \eqref{eq:noisy_or}.
Equivalently, if $\tilde{c}_{ij}(t_k)$ is interpreted as the independent detection probability of satellite $i$ at target $j$, then
\begin{equation}
    \Pr\!\left[\bigcup_i \text{sat } i \text{ covers } j\right]
    = 1 - \prod_i \Pr\!\left[\text{sat } i \text{ misses } j\right]
    = 1 - \prod_i (1 - \tilde{c}_{ij}(t_k)).
    \label{eq:noisy_or_derivation}
\end{equation}
No separate relaxation is needed: the noisy-OR operator is the unique member of the probabilistic t-conorm family that is smooth, bounded in $[0,1]$, and strictly increasing in each argument.

\subsection{R4: Mellowmax $\Leftrightarrow$ LogSumExp + constant}

The mellowmax operator is introduced by \citet{asadi_alternative_2017} as an alternative to the Boltzmann softmax used for value aggregation in $Q$-learning, where it restores a non-expansion property Boltzmann softmax lacks. Mellowmax is equivalent to LogSumExp up to an additive constant:
\begin{equation}
    \operatorname{mm}_{\omega}(X)
    \;=\; \frac{1}{\omega}\log\!\left(\frac{1}{K}\sum_{k=1}^{K} e^{\omega x_k}\right).
    \label{eq:mellowmax}
\end{equation}
Setting $\omega = 1/\beta$ and expanding the $1/K$ factor gives
\begin{equation}
    \operatorname{mm}_{1/\beta}(X)
    = \beta \log\!\left(\sum_k e^{x_k/\beta}\right) - \beta \log K
    = \logsumexp_\beta(X) - \beta \log K,
    \label{eq:mellowmax_to_lse}
\end{equation}
so the two differ by the constant $\beta \log K$. Hence, the optimizer sees identical dynamics under mellowmax.

\end{document}